\documentclass[runningheads]{llncs}

\usepackage[mobile]{eccv}

\usepackage{multirow}
\usepackage{algpseudocode}
\usepackage{array}
\newcolumntype{C}[1]{>{\centering\let\newline\\\arraybackslash\hspace{0pt}}m{#1}}
\usepackage{algorithm}
\algnewcommand{\LineComment}[1]{\State \(\triangleright\) #1}

\usepackage{eccvabbrv}

\usepackage{graphicx}
\usepackage{booktabs}

\usepackage[accsupp]{axessibility}  

\usepackage{hyperref}

\usepackage{orcidlink}

\begin{document}

\title{Robustifying Point Cloud Networks by Refocusing} 


\author{Meir Yossef Levi\inst{1} \and
Guy Gilboa\inst{1}}

\authorrunning{Meir Yossef Levi and Guy Gilboa}

\institute{Technion - Israel Institute of Technology, Haifa, Israel \\
\email{me.levi@campus.technion.ac.il}\\
\email{guy.gilboa@ee.technion.ac.il}}

\maketitle

\begin{abstract}
  The ability to cope with out-of-distribution (OOD) corruptions and adversarial attacks is crucial in real-world safety-demanding applications. In this study, we develop a general mechanism to increase neural network robustness based on focus analysis.
  Recent studies have revealed the phenomenon of \textit{Overfocusing}, which leads to a performance drop. When the network is primarily influenced by small input regions, it becomes less robust and prone to misclassify under noise and corruptions.
  However, quantifying overfocusing is still vague and lacks clear definitions. Here, we provide a mathematical definition of \textbf{focus}, \textbf{overfocusing} and \textbf{underfocusing}. The notions are general, but in this study, we specifically investigate the case of 3D point clouds.
  We observe that corrupted sets result in a biased focus distribution compared to the clean training set.
  We show that as focus distribution deviates from the one learned in the training phase - classification performance deteriorates.
  We thus propose a parameter-free \textbf{refocusing} algorithm that aims to unify all corruptions under the same distribution.
  We validate our findings on a 3D zero-shot classification task, achieving SOTA in robust 3D classification on ModelNet-C dataset, and in adversarial defense against Shape-Invariant attack. Code is available in:
  
  \href{https://github.com/yossilevii100/refocusing}{https://github.com/yossilevii100/refocusing}.
\end{abstract}

\section{Introduction}
\label{sec:intro}

In recent years, significant research efforts have been dedicated to understanding the underlying mechanisms of neural networks for producing accurate and reliable results. One would like to understand where the network "looks" within an image to establish classification, a field known as explainable AI (XAI) \cite{gradcam,lrp,transformer_explainable_ai}. It has already been revealed that networks may rely on spurious cues \cite{natural_adversarial_examples} or shortcuts \cite{learning_shortcuts} for prediction. For instance, a cow may be predicted based on a green pasture rather than cow features, influenced by dataset bias. XAI algorithms typically propagate gradient computations to provide a heatmap, showing regions in the image which mostly contributed to the network's decision. XAI assists in the debugging process of neural network architecture development.

Furthermore, XAI maps can offer new insights into the network's behavior for a given input, based on focus analysis.
\begin{figure*}[ptbh!]
  \centering
   \includegraphics[width = 0.9\linewidth, height =0.45 
   \linewidth]{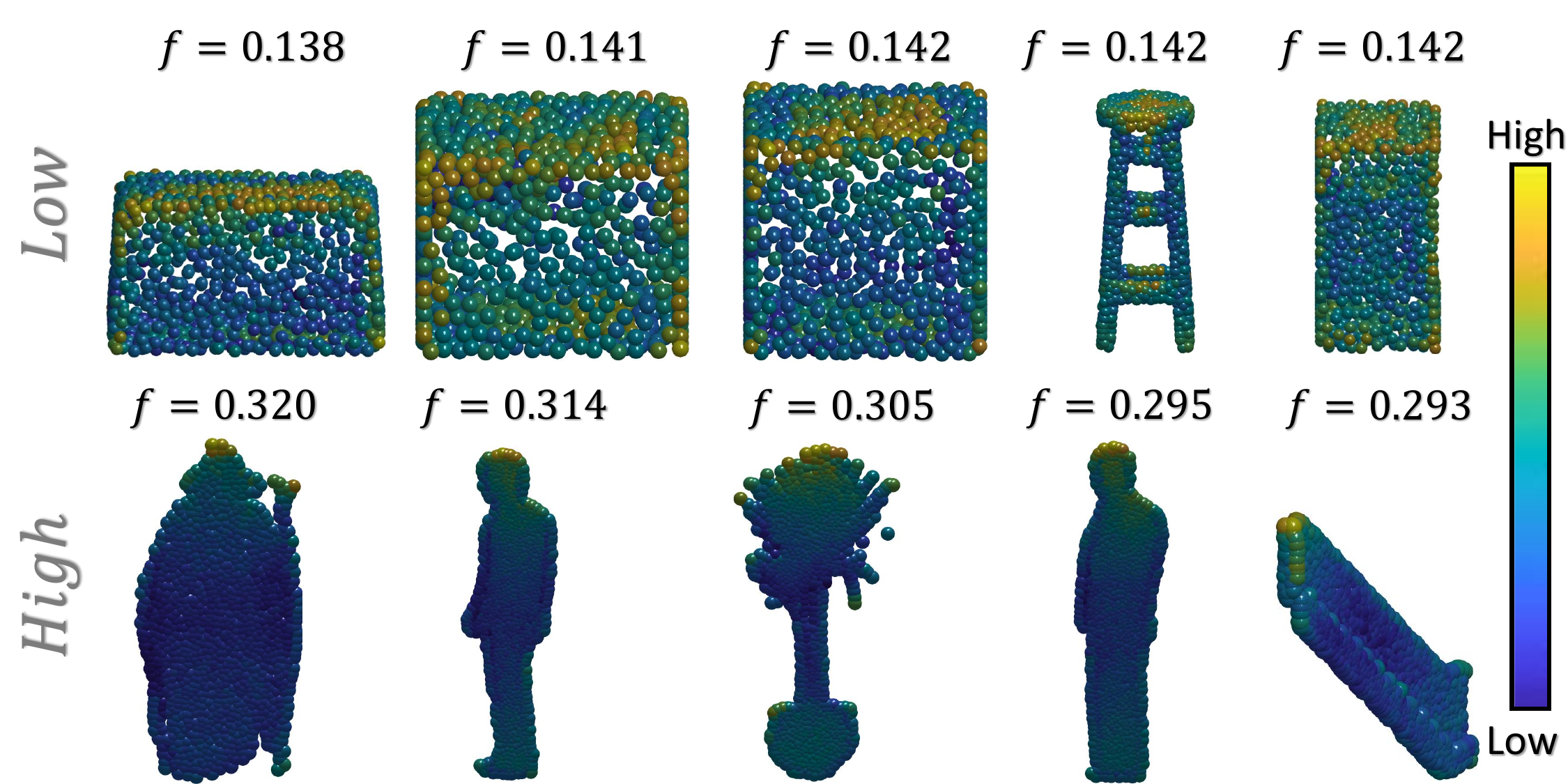}
   \caption{{\bf High and low focus examples.} Samples resulting in low focus distribute influence across a broad spectrum of points, while in the case of high focus, influence becomes concentrated within specific regions. Samples containing flat areas, with some extent of symmetry against the center of the shape, contribute to a decrease in focus. It is possible that samples predominantly spanning a 2D plane prompt the network to prioritize attention towards distinctive regions. Points are color-coded by influence.}
   \label{fig:low_to_high_clean}
\end{figure*}
By measuring the influential regions (using XAI tools), it is possible to investigate the relations between the \emph{focus} of the network and its performance. Although a definition of attention concentration based on entropy was introduced in \cite{What_does_attention_pay_attention_to}, the term focus, in a general context, remains somewhat ambiguous and not well-defined. Intuitively, a focused network is influenced by only a few prominent input data points, while an unfocused one relies on input data spread throughout the domain. See \cref{fig:low_to_high_clean} for point cloud examples of high and low focus.

Recent studies have shown that many networks tend to overfocus, making decisions based only on a few highly localized input regions \cite{robustifying_token_attention}. This results in less stable performance and lacks robustness when statistics change with respect to the training phase.
We investigate classification robustness in the context of 3D point clouds, examining the interplay between focus and robustness to corruptions and to adversarial attacks. 

\begin{figure*}[htbp!]
  \centering
   \includegraphics[width=0.9\linewidth, height=0.507\linewidth]{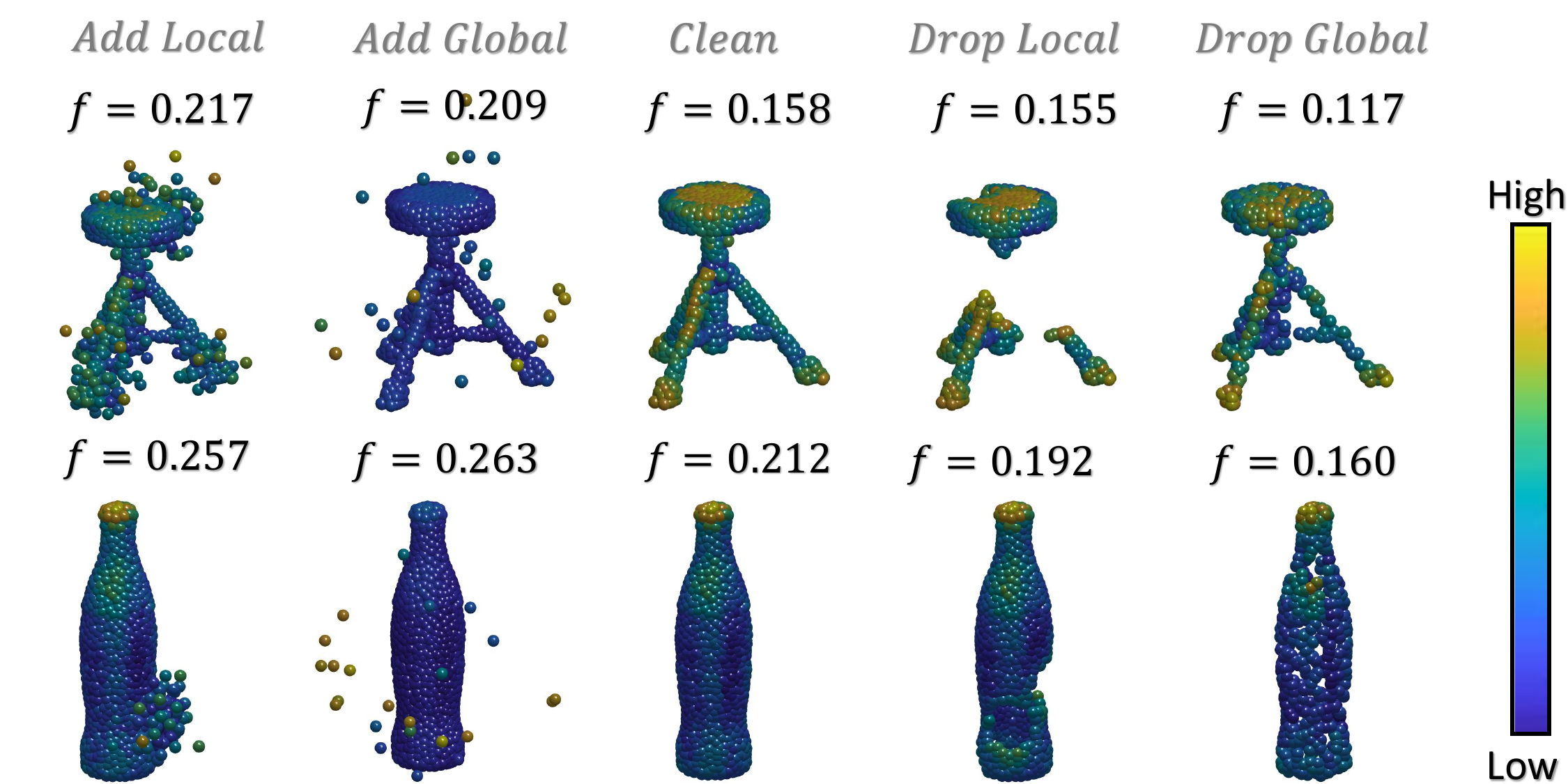}
   \caption{{\bf Variation of focus across different corruptions.}  Influence maps on corrupted samples from ModelNet-C \cite{modelnet_c} using DGCNN \cite{dgcnn}. The presence of outliers predominantly increases focus, while occluded parts decrease it.
   Points are color-coded according to influence.
   We denote by $f$ the focus of the network for that sample, as defined in Eq. \eqref{eq:focus}.
   }
   \label{fig:low_to_high_corruptions}
\end{figure*}

We first define a general notion of the network's focus based on normalized entropy.
We then analyze the focus distribution and its changes under various corruptions, using a recently proposed corrupted point cloud benchmark, ModelNet-C \cite{modelnet_c}. Our findings reveal that the clean dataset, used during training, has a distinct focus distribution for which the network is optimal. Each corruption type induces a different unique focus distribution. The general trend is that corruptions involving outliers cause overfocus, while those involving occluded parts cause underfocus, compared to the uncorrupted data (See \cref{fig:low_to_high_corruptions}). This leads to significant performance degradations.

We propose a new learning procedure that reduces the variance of the focus distribution under corruptions. In a nutshell, we train the network to perform a more challenging task, relying on less influential input. This results in a less focused network during training. Subsequently, by applying the same filtering during inference, which may contain corruptions, we achieve a focus distribution which is more aligned to the training phase.
A more stable network is obtained, with improved robustness to out-of-distribution (OOD) corruptions, effectively balancing overall performance. This generic idea can enhance the robustness of various point cloud classification networks (we demonstrate it on DGCNN \cite{dgcnn}, RPC \cite{modelnet_c}, and GDANet \cite{gdanet}).

\subsection{Why rely on less influential inputs?}

\begin{figure}
  \centering
   \includegraphics[width=0.8\linewidth, height=0.5\linewidth]{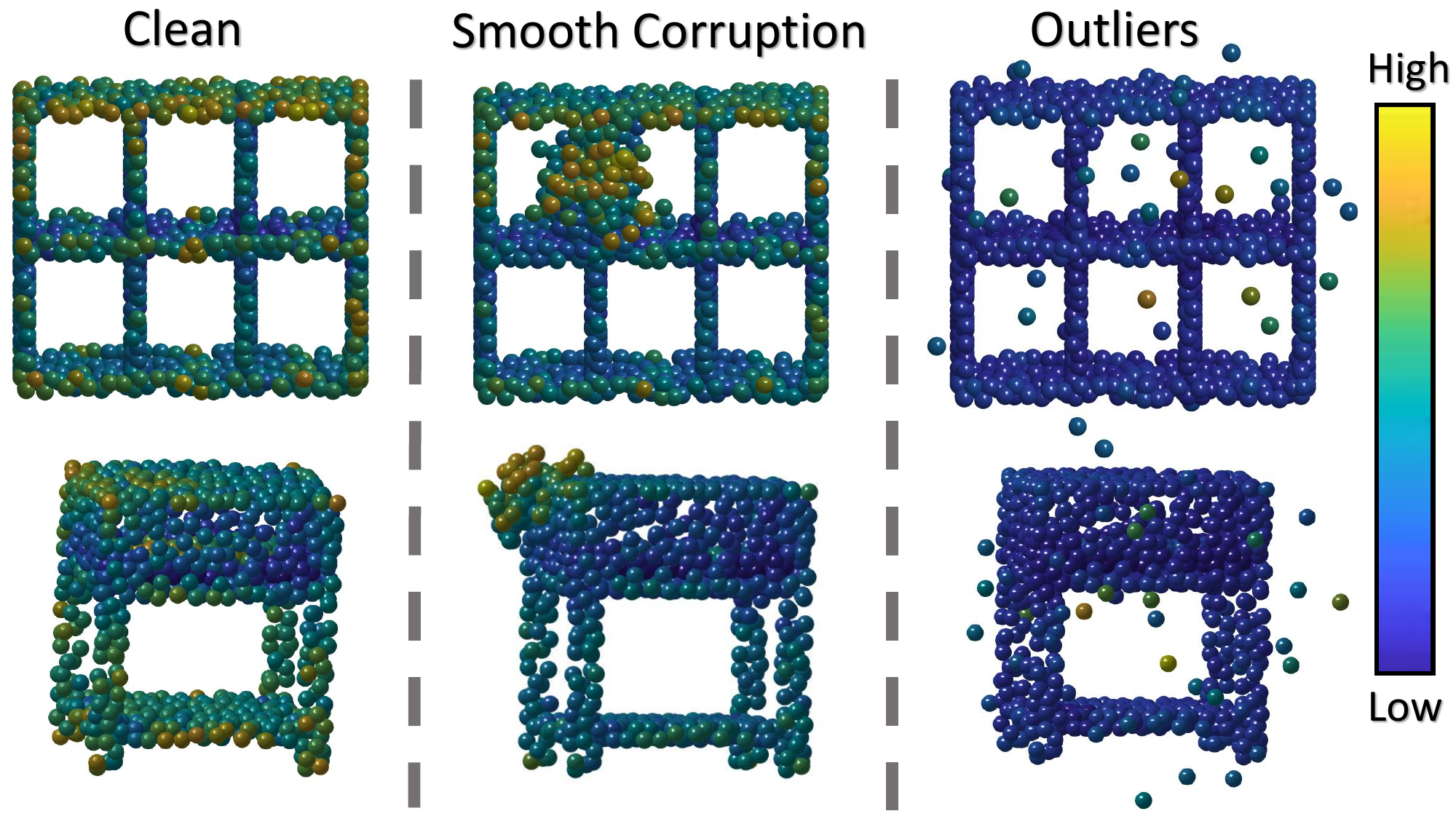}
   \caption{\textbf{Outliers and smooth corruptions often draw high influence.} Samples color coded according to influence. The measurement of influence, reasonably distributed in an uncorrupted point cloud, is mainly redistributed to outliers in the corrupted version.}
   \label{fig:Outliers_influence}
\end{figure}

\begin{figure*}[htbp!]
  \centering
   \includegraphics[width=1\linewidth, height=0.4\linewidth]{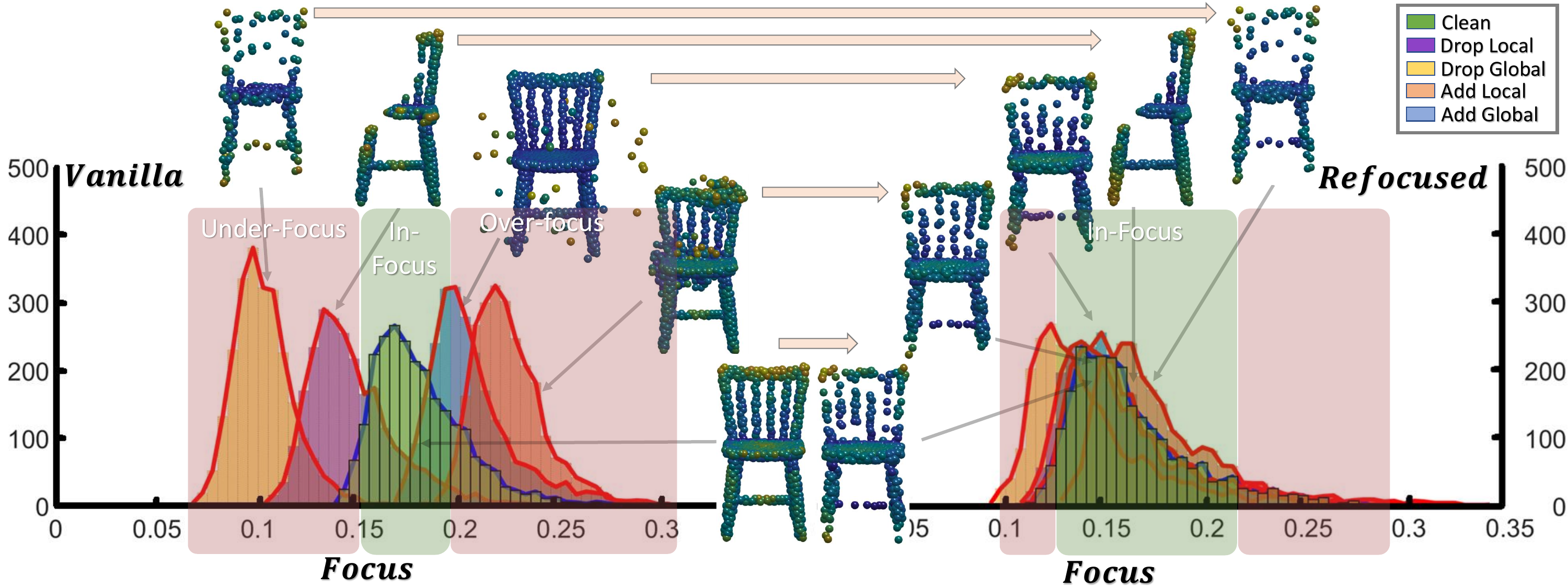}
   \caption{\textbf{Refocusing.} Left - Focus distribution on DGCNN \cite{dgcnn} based on the clean set of ModelNet40 \cite{modelnet40} and on corrupted sets (ModelNet-C \cite{modelnet_c}). Corrupted samples deviate from the in-focus region. Right - Screening out influential points aligns the focus distribution, expanding the in-focus region at the expense of under- and over-focus regions. In the chair example shown, one can observe the network is influenced by similar points after refocusing, resembling roughly the same influence distribution across the shape.}
   \label{fig:refocus}
\end{figure*}

Our approach might seem counterintuitive, as one would expect prominent regions to contribute most to high-quality class discrimination.
However, relying on less influential points offers several important benefits:

\begin{enumerate}
    \item \textbf{Calibrated focus.} Our analysis demonstrates that the network performs optimally for data within the focus distribution of the training phase (\cref{fig:histogram_and_success_rate}). The proposed learning procedure of refocusing by screening out the influential points is highly stable, yielding a similar focus distribution for OOD samples as for the clean set, as illustrated in \cref{fig:refocus}.
    
    \item \textbf{Better resilience to corruptions.} Corruptions are often perceived as influential features by the network, leading to a significant performance drop, as shown in Fig. \ref{fig:Outliers_influence}. Our approach exhibits implicit significant filtering capacity of outliers. In \cref{table:most_vs_least}, we conducted a comparison between filtering the most influential and filtering the less influential points to analyze the trade-off between accuracy on the clean set and on the corrupted set. We find that relying on the less influential points significantly enhances robustness to outliers at the expense of only a slight performance drop on the clean set.
    
    \item \textbf{Preserving clean data performance.} Sub-sampling, known for enhancing robustness \cite{epic, pointguard}, may cause a slight performance drop. However, this can be compensated through ensemble methods \cite{epic}, which, overall, can surpass vanilla performance, as validated in \cref{table:robust_classification}.
\end{enumerate}

\begin{table}[htbp]
\centering
\begin{tabular}{p{2.5cm} || C{1cm} C{2.5cm}}
\hline
\multirow{2}{*}{} &
\multicolumn{2}{c}{Accuracy} \\
\cmidrule(l){2-3}
& Clean & Add-Global \cite{modelnet_c} \\
\hline
Least Influential & 91.0\% & 90.3\% \\
Most Influential & 91.1\% & \textbf{43.2\%} \\
\hline
\end{tabular}
\caption{\textbf{Test accuracy of most vs. least influential inputs.} While both approaches perform comparably on the clean set, using less influential points is much more robust to outliers. Trained and evaluated on 600 least or most influential.}
\label{table:most_vs_least}
\end{table}

Our main contributions are:
\begin{enumerate}
    \item We provide a general definition for the network's focus and for over- and under-focusing. A comprehensive analysis is performed relating corruptions to the focus distribution.
    \item We propose a refocusing scheme, which is parameter-free and can be applied to any point-cloud classification network. It offers a more robust and reliable learning strategy, handling OOD corruptions and attacks better, without sacrificing overall accuracy.
    \item We demonstrate our approach in \emph{Robust Classification} and \emph{Adversarial Defense}, achieving State-of-the-Art (SOTA) results.
\end{enumerate}

\section{Related Work}
\textbf{Focus in neural networks.} Focus has been investigated until now mainly in the context of attention maps of transformers. The studies of \cite{attention_and_convolution} and \cite{pay_less_attention} have highlighted the tendency of attention maps in certain network layers to rely heavily on a few dominant tokens. Guo et al. \cite{robustifying_token_attention} coined this phenomenon as \textit{Token Overfocusing} and established a correlation with corruptions. While there have been visual demonstrations of overfocused attention maps, we lack a clear and standardized definition of focus, which may be applied to any type of network.
Ghader et al. \cite{What_does_attention_pay_attention_to} introduced \textit{Attention Concentration} defined mathematically using attention entropy. This definition has allowed for the analysis of attention concentration in different parts of sentences in natural language processing (NLP) \cite{Analyzing_the_structure_of_attention}, and the investigation of differences between supervised and unsupervised training in terms of attention entropy \cite{explanation_on_pretraining_bias}. However, it is primarily designed for attention maps in transformers and cannot be applied to general neural network architectures. Additionally, it does not rely on normalized entropy, which plays a vital role when dealing with varying numbers of input elements.

\textbf{3D robust classification.}
Point cloud classification \cite{dgcnn, curvenet, cloud_walker, pointnet, pointnet++, point_mlp, pointbert, gdanet, paconv, pct} is vital for autonomous driving \cite{autonomous_driving1, autonomous_driving2} and robotics \cite{robotics}. However, research on robustness against corruptions is relatively scarce.
PointNet \cite{pointnet} introduced the concept of \textit{critical points}, which is a subset of points that remain active after the last pooling layer.
We note that outliers are often misinterpreted as influential or critical.
Supervised \cite{learning_to_sample, samplenet} and unsupervised \cite{self_ordering_point_clouds} 3D sorting strategies have been proposed to better sample point clouds for downstream tasks. These approaches prone to underperform at the presence of out-of-distribution (OOD) corruptions since they prioritize highly influential points. Our observations indicate outliers are highly likely to be sampled by these methods.
Several studies \cite{pointcleannet, point_asnl} offered learnable outlier removal for adaptive sampling in Euclidean space. 
ModelNet-C dataset \cite{modelnet_c} introduced real-world corruptions, involving outliers or missing points (which can be caused by occlusions) from 3D point clouds, either globally or locally. They also proposed Robust Point-cloud Classifier (RPC) \cite{modelnet_c}, an algorithm which is a combination of the most robust modules from typical classification networks, achieving state-of-the-art performance on ModelNet-C. Recently, EPiC \cite{epic} proposed an ensemble approach combining different sampling schemes, outperforming RPC. However, such ensemble methods are relatively slow and resource-intensive.

\textbf{3D adversarial attacks.}
Designing classification networks which are robust  against adversarial attacks, particularly in 3D settings, is significant. Numerous 3D adversarial attack methods have emerged in recent years \cite{minimal_adversarial, adversarial_shape_perturbations, shape_adv, robust_adversarial_objects, saliency_maps, adversarial_attack_and_defense, explainable_one_point_attack, Generating_3d_adversarial_point_clouds, advpc, lg_gan, isometry_attack, deep_learning_in_an_adversarial_setting}.
These attacks primarily focus on perturbing points, emphasizing imperceptible manipulations. Our proposed influence measure and point filtering approach can be employed for adversarial defense. Shape-Invariant Attack \cite{shape_invariant} introduces a sensitivity map consistent across diverse neural networks, sliding points along the tangent plane based on this map.
Point Cloud Saliency Maps \cite{saliency_maps} analyze gradient loss when shifting points to the spherical center to determine importance.
A line of attacks \cite{adversarial_attack_and_defense, deep_learning_in_an_adversarial_setting, Generating_3d_adversarial_point_clouds} is initiated from critical points \cite{pointnet} as vulnerabilities of the network.
We compare our proposed defensive scheme against the state-of-the-art Shape-Invariant Attack \cite{shape_invariant}, highlighting that even imperceptible perturbations can alter point influence, demonstrating the generality of our approach.

\textbf{3D adversarial defense.}
Advanced 3D augmentation techniques like PointWolf \cite{pointwolf} and RSMix \cite{rsmix} enhance network robustness against corruptions \cite{modelnet_c}. Adversarial Training (AT) techniques \cite{Ensemble_adversarial_training, Pointcutmix, extending_adversarial_attacks_and_defenses, adversarial_attack_and_defense, pagn, lpf_defense, point_acl} intentionally introduce perturbations during training to defend against malicious attacks, but they require prior knowledge, therefore not robust for OOD corruptions.
PointGuard \cite{pointguard} and PointCert \cite{point_cert} propose certified defense schemes using point-cloud sub-sampling and majority voting. However, their ensemble strategy is highly demanding computationally.
Point filtering techniques include 
Simple Random Sampling (SRS) \cite{dupnet},
which removes input points randomly, and Statistical Outlier Removal (SOR) \cite{sor}, which filters points far from their nearest neighbors. 
SOR performs well against outliers (with known distributions) but may fail on smooth corruptions. Dup-Net \cite{dupnet} 
combines denoising and upsampling, significantly increasing inference time.
LPF-Defense \cite{lpf_defense} focuses on low-frequency features using spherical harmonics transformation. IF-Defense \cite{if_defense} optimizes the point-cloud in terms of surface distortion, requiring a training phase during inference. In
\cite{extending_adversarial_attacks_and_defenses} prediction derivatives are calculated to obtain per-point importance, 
facing scalability challenges with large networks.
Our work aims to address these challenges. In our comparison we use Shape-Invariant attack \cite{shape_invariant} and LPF-Defense \cite{lpf_defense} as baselines.

\section{Focus Definition and Refocusing Method}
Let $X \in \mathbb{R}^{N \times d}$ be the input data consisting of $N$ elements $X_i \in \mathbb{R}^{d}$ in  arbitrary dimension $d$.
Let $F:\mathbb{R}^{N \times d}\rightarrow \mathbb{R}^m$ be a  neural network. 
We denote by $I_{F}(X) : \mathbb{R}^{N \times d} \rightarrow \mathbb{R}^N$ an influence score. This measure attempts to quantify, for a given $X$, the amount of influence of each input element on the output of the network $F$. The term \emph{influence} may be defined in various ways with multiple approaches to compute it, such as based on attention mechanisms or in the general case by using XAI methods. We propose below a simple and fast influence measure for point cloud networks.
We denote by $I_{F}^i$ the $i^{th}$ element of $I_F(X)$. It is further assumed that the
influence score is normalized, such that it has only non-negative values with a unit sum, that is,
 $I_F^i \ge 0$, $\forall i = \{1,..\,\,,N\}$, and  $\sum_{i=1}^N I_F^i = 1$.  
The influence score is used in this study to define a general notion of the network's focus. 

\subsection{Focus}
 Let $p$ be a distribution of $N$ elements with $p_i$ denoting the probability of each element. We remind that $p_i \ge 0$, $\forall i$, and  $\sum_{i=1}^N p_i = 1$.
Given some probability distribution $p$, a general measure for uniformity of that distribution is entropy,
\begin{equation}
    H(p) := -\sum_{i=1}^{N}{p_i \ln (p_i) }.
  \label{eq:H}
\end{equation}
It is a non negative function, low when the distribution has sharp peaks and increases as the distribution becomes more even. To obtain a normalized measure, in the range $[0,1]$, we use normalized entropy. This measure divides the entropy by the maximum possible entropy for that sample: 
$H_n := H / H_{max}$, where $H_{max} = \max_{p} \{H(p)\}$. It is well known that entropy is maximized for the uniform distribution, $p_i=1/N$, $\forall i$. Plugging this in $H_{max}$ and rearranging yields the following expression for 
the normalized entropy \cite{normalized_entropy},
\begin{equation}
    H_n(p) = \frac{H(p)}{\ln(N)}.
  \label{eq:H_n}
\end{equation}
We can now define the focus of a network. 
\begin{definition}[Focus of a network]
\label{def:focus}
Given a network $F$, an input $X$ and an associated normalized influence measure $I_F(X)$, the focus of the network, denoted $f$, is defined by 
\begin{equation}
    f(X) := 1 - H_n(I_F(X)).
  \label{eq:focus}
\end{equation}
\end{definition}
Let us state some basic properties of $f(X)$.

\begin{proposition}[Focus properties]
\label{proposition_1}
For any network $F$, input $X$ of any size $N$ and normalized influence  $I_F(X)$, the focus $f(X)$ has the following properties: 
\begin{enumerate}
    \item $f(X) \in [0,1]$.
    \item $f(X)=1$ iff $\exists i$,  $I_F^i=1$, $I_F^j=0$, $\forall j\ne i$.
    \item $f(X)=0$ iff  $I_F^i=\frac{1}{N}$, $\forall i$.
\end{enumerate}
\end{proposition}
The proof follows directly from the properties of normalized entropy. Consequently, we obtain a general definition of focus within the range $[0,1]$, enabling easy comparison across different network settings and input sizes. To provide a better intuition for the extremities of the focus measure distributed on a 3D shape, we visualize samples from ModelNet40 \cite{modelnet40} with high and low focus values in \cref{fig:low_to_high_clean}.

\subsubsection{Why normalized entropy?}

As mentioned in \cite{normalized_entropy}, "To obtain a measure of uncertainty that can be compared across distributions, actual uncertainty must be divided by the maximum possible uncertainty." In this paper, our primary focus is on 3D classification; however, we formulate the definition in a broader context, such that it can be extended to other domains.
For an entropy measure, the following relation holds: 
\[-\sum_1^{k_1}\frac{1}{k_1} \ln\left(\frac{1}{k_1}\right) > -\sum_1^{k_2}\frac{1}{k_2} \ln\left(\frac{1}{k_2}\right), \text{ for all } k_1 > k_2 > 0.\]
This implies that the maximal entropy of a vector with more elements is higher. In 3D analysis, when the influence measure is evenly distributed, a larger point cloud has larger entropy. We would like a measure which is invariant to the point cloud size. 
Hence, normalized entropy is employed.

\subsection{Focus distribution, over- and under-focusing}
We would like to analyze the network's focus behavior under different datasets, introducing additional notions.
Let $T \in \mathbb{R}^{M_T \times N \times d}$ represent a training set with $M_T$ data instances, and $S \in \mathbb{R}^{M_S \times N \times d}$ denote a test set with $M_S$ data instances. Random samples from these sets are denoted as $X^T$ and $X^S$, respectively. The empirical mean is denoted by $E[\cdot]$.
For a given set $Q$, let $\mu_Q = E[f(X^Q)]$ and $\sigma_Q = \sqrt{E[(f(X^Q)-\mu_Q)^2]}$.

\begin{figure*}[ptbh!]
  \centering
   \includegraphics[width=1\linewidth, height=0.5\linewidth]{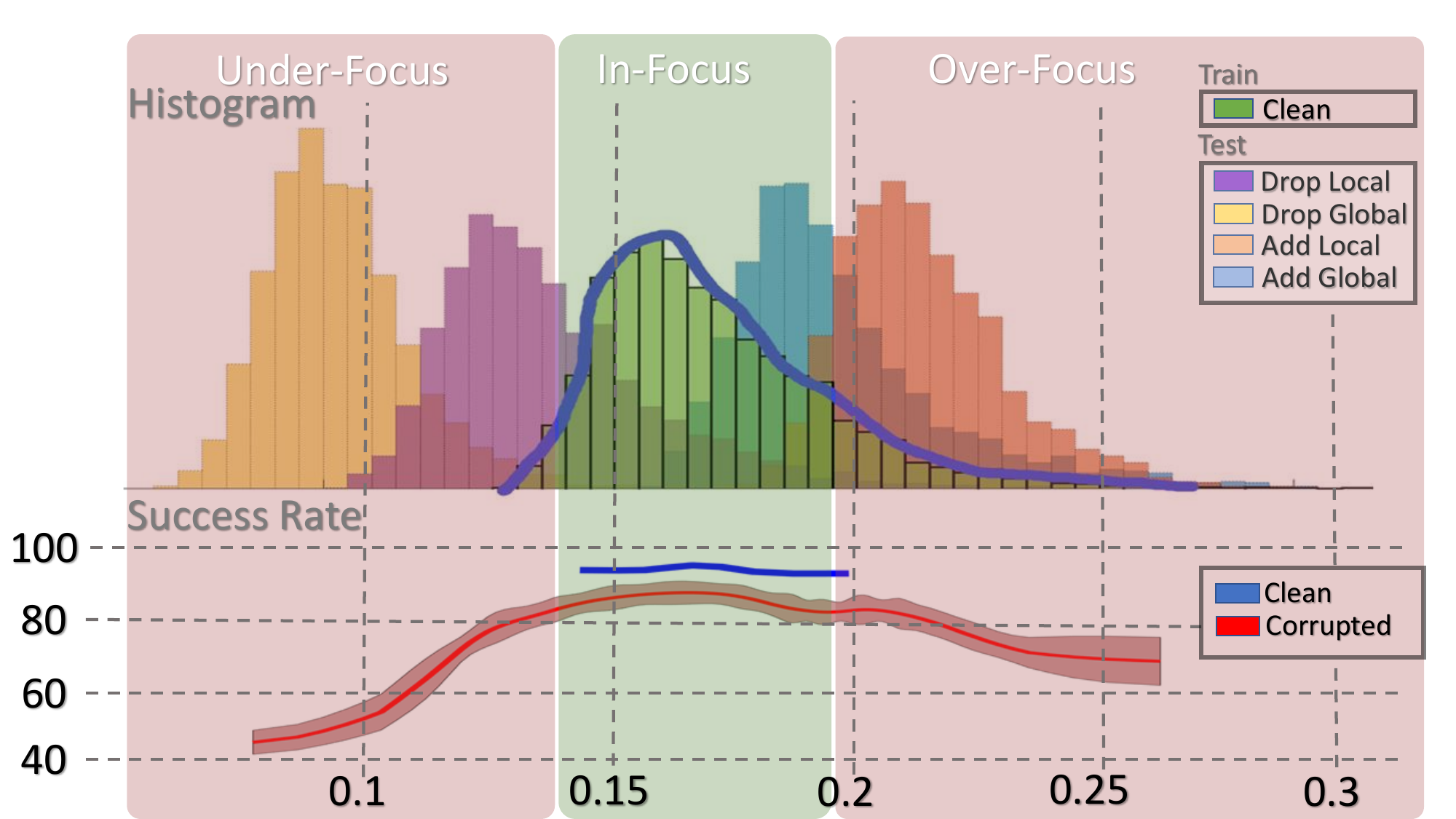}
   \caption{\textbf{In-focus, Under-focus, and Over-focus.} Top - Histogram of focus values for the clean set, ModelNet40 \cite{modelnet40}, defining the in-focus region inside the standard deviation. Histograms collected from corrupted sets in ModelNet-C \cite{modelnet_c} are clearly out of the training distribution. The trend indicates that the appearance of outliers correlates with over-focusing, while the absence of points correlates with under-focusing. Bottom - Success rate of clean (blue) and corrupted (red) sets. A clear performance drop is observed in the over-focus and under-focus regions.}
   \label{fig:histogram_and_success_rate}
\end{figure*}

We define an over-focused sample as $X^S$ such that $f(X^S) \ge \mu_T + \alpha\cdot\sigma_T$ and an under-focused sample as $X^S$ such that $f(X^S) \le \mu_T - \beta\cdot\sigma_T$, where $\alpha$ and $\beta$ are tunable parameters corresponding to $T$.
Practically, this approach allows us to identify OOD regions, where samples significantly deviate from the learned focus distribution. Detecting such regions is crucial, as performance degradation is associated with data residing outside the training distribution (See \cref{fig:histogram_and_success_rate}). By incorporating prior knowledge derived from the training statistics, we gain valuable insights into potential challenges the model may encounter when faced with unfamiliar data.

\subsection{Refocusing}

Our method relies on filtering the most influential points identified by a given influence map, as outlined in \cref{alg:inference}. To achieve this, we seek an influence evaluation that is computationally efficient, given its integration during inference. The literature on XAI for point-cloud classification broadly falls into three categories: 1) Iterative processes \cite{point_lime, saliency_maps}; 2) Dedicated explainable architectures \cite{pointhop, xpcc}; 3) Utilizing gradients \cite{gradient_based, bubblex}. However, these approaches either consume significant computational time or lack the capacity to explain certain architectures. Consequently, we choose an explainable method that is a variation of \cite{self_ordering_point_clouds}.
In essence, the influence measure prioritizes importance based on the frequency of appearance in the global feature vector. Specifically, it quantifies the count of features with the highest values compared to all other points. The influence measure is defined as:

\begin{equation}
    I_F(j) = \sum_{k=1}^\mathcal{K} \mathbb{I}(j==\arg\max_{n}(X_f(n,k))),
  \label{eq:importance}
\end{equation}
where $\mathbb{I}$ is an indicator function (equal to 1 when true and 0 otherwise), and $\mathcal{K}$ is the size of the feature vector.

\begin{figure}[ptbh!]
    \centering
    \captionsetup[subfigure]{justification=centering}
    \begin{subfigure}[t]{0.48\linewidth} 
\includegraphics[width=1\linewidth]{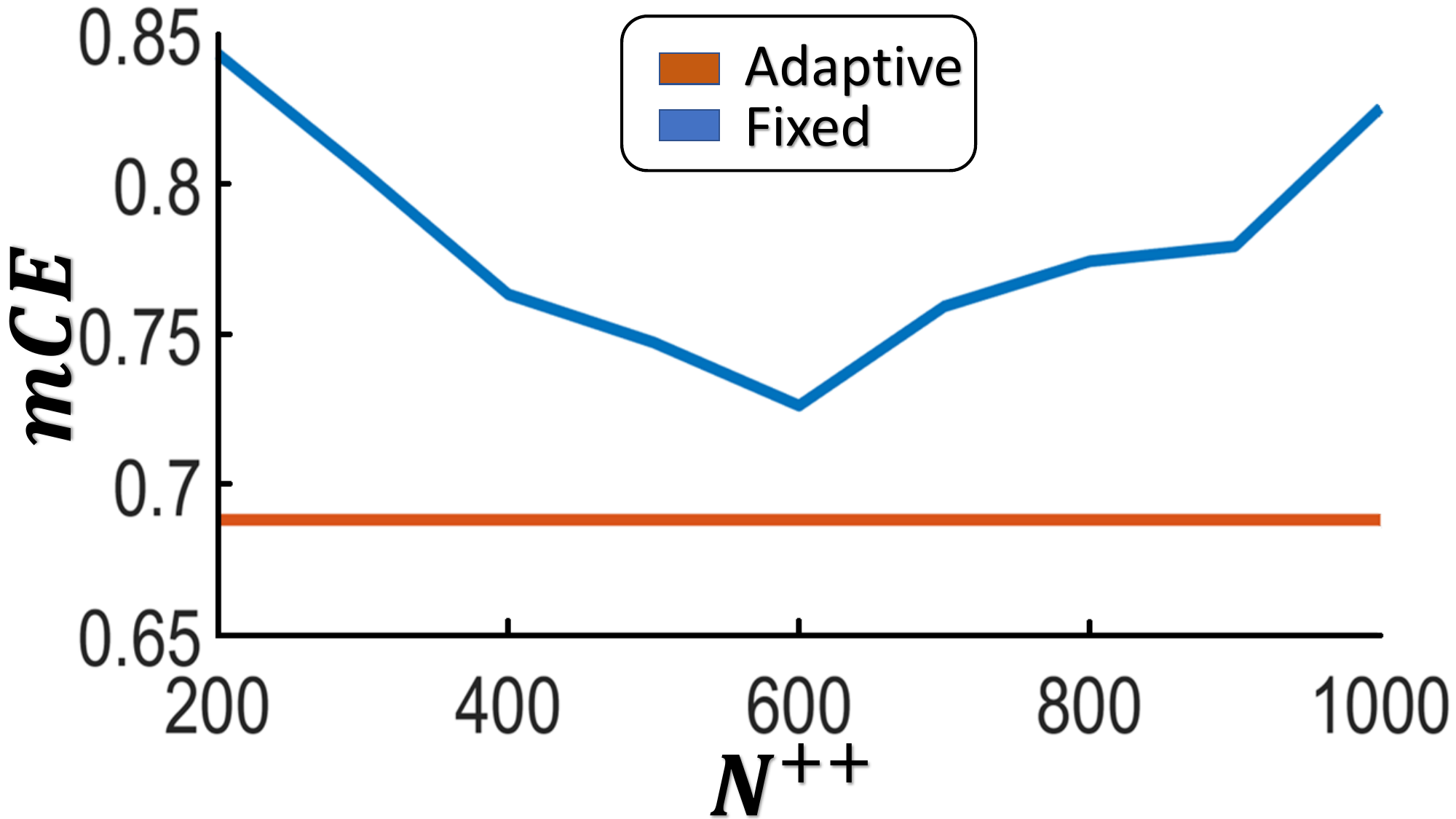}
\caption{{\bf  Mean Corruption Error (mCE) Vs. Number of retained points.}}
        \label{subfig:ablation_mce}
    \end{subfigure}
    \begin{subfigure}[t]{0.48\linewidth} 
\includegraphics[width=1\linewidth]{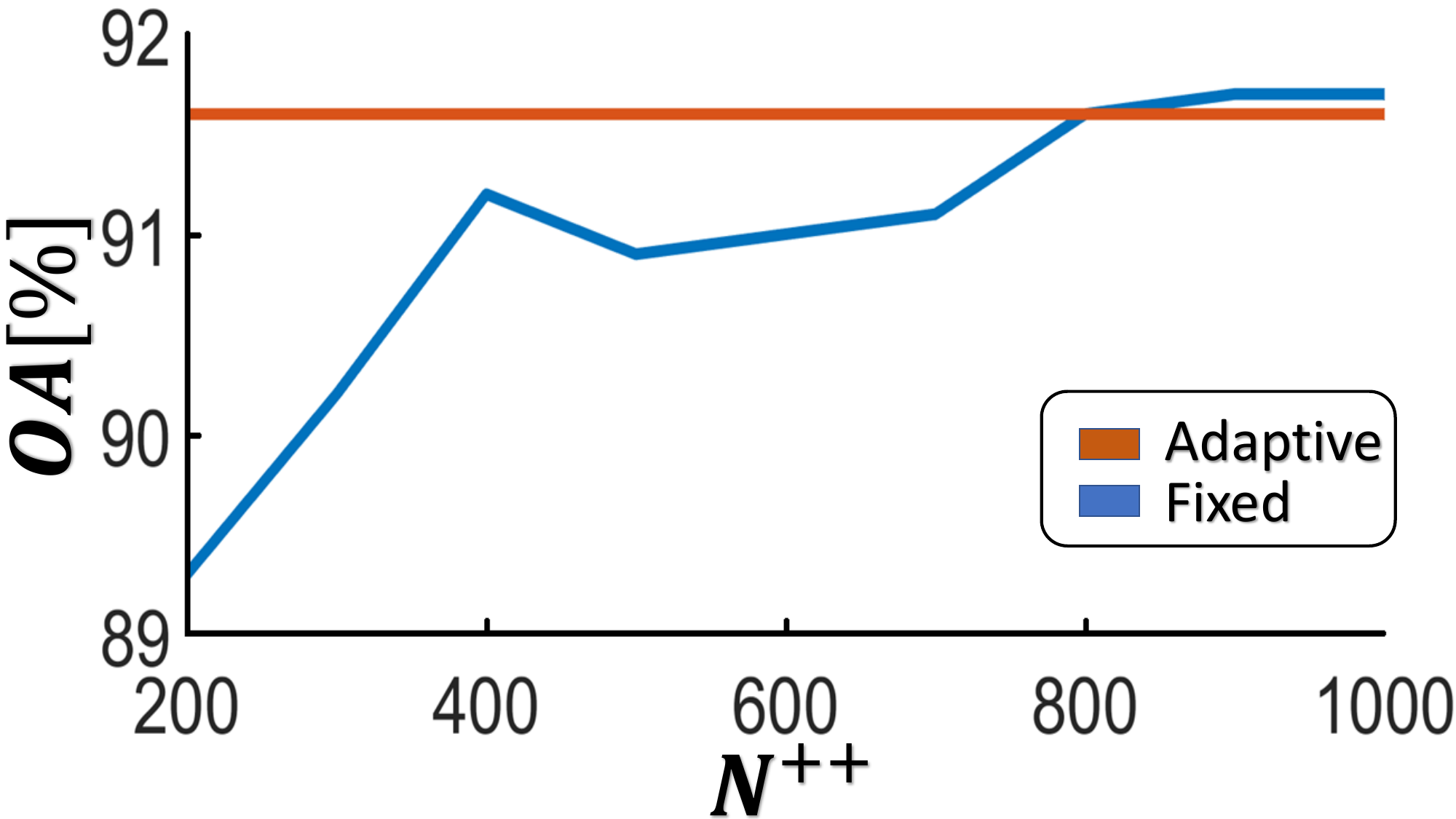}
\caption{{\bf Overall Accuracy (OA) Vs. Number of retained points.} }    
        \label{subfig:ablation_oa}
    \end{subfigure}
    \caption{{\bf Adaptive vs. fixed threshold.} Our adaptive threshold demonstrates superior robustness compared to any fixed threshold. The adaptive, parameter-free threshold yields a single result, represented as a straight (orange) line in the plot.}
    \label{fig:adaptive_vs_fixed}
\end{figure}

\subsubsection{Refocusing - ``reign of the less influential''}
In Fig. \ref{fig:histogram_and_success_rate} the success rate is shown as a function of focus.
We see typical narrow range for the clean set and a much wider range for the corrupted sets.
Based on our observation that outliers strongly affect the influence map, it is intuitive to diminish corruption byproducts by discarding the most influential points. Thus, the influence is redistributed among the remaining points. This action should filter out corruptions and align the focus closer to the narrow region of the clean set. Introduction or removal of points are shifting the focus distribution from the distribution learned during training. Outliers cause over-focusing, whereas missing points yield under-focusing. However, after cropping most influential points, the focus distribution is aligned, as can be seen in Fig. \ref{fig:refocus}. Therefore, we term our process as \textit{refocusing}. 

\begin{algorithm}
\caption{Refocusing (Inference)}\label{alg:inference}
\begin{algorithmic}
\Require{$X, params_{refocused}$} 
\State $model_{refocused} \gets params_{refocused}$\Comment{Load Pretrained}
\State $X_f = model_{refocused}(X)$ \Comment{First forward-pass}
\State $I(X) = Eq. \eqref{eq:importance}$ 
\State $\hat{I} = \frac{I}{\Sigma_{i=1}^N(I)}$
\State $f = Eq. \eqref{eq:focus}$ \Comment{Calculate $f$}
\State $K = \lfloor (1-f) \cdot N \rfloor$  \Comment{Adaptive Threshold}
\LineComment{Select K lowest influential points}
\State $X_{sampled} = SelectLowest(X, \hat{I}, K)$
\State $P = model_{refocused}(X_{sampled})$ \Comment{Second forward-pass}
\State $Class = \arg\max(P)$
\end{algorithmic}
\end{algorithm}

\subsubsection{Adaptive threshold}
We argue that samples containing outliers should be subjected to more aggressive filtering, compared to samples that have missing points. In fact, it is not clear whether the latter case should be sampled at all. This raises the general question: \textit{How many points should be retained?}
In information theory, \emph{normalized entropy} \cite{normalized_entropy}, also referred to as \emph{efficiency} can resolve this. One can think of maximal entropy as the most efficient representation, where normalized entropy is a measure of relative efficiency. In the context of 3D classification, setting equal contribution for any input point is equivalent to the most efficient representation. Thus, we advocate using normalized entropy as a criterion for determining the ratio of remaining points during the filtering process. We set the remaining number of points to $N^{++} = \lfloor({1-f})\cdot N\rfloor$.

\begin{figure}[ptbh!]
  \centering
   \includegraphics[width = 0.8\linewidth, height =0.5 
   \linewidth]{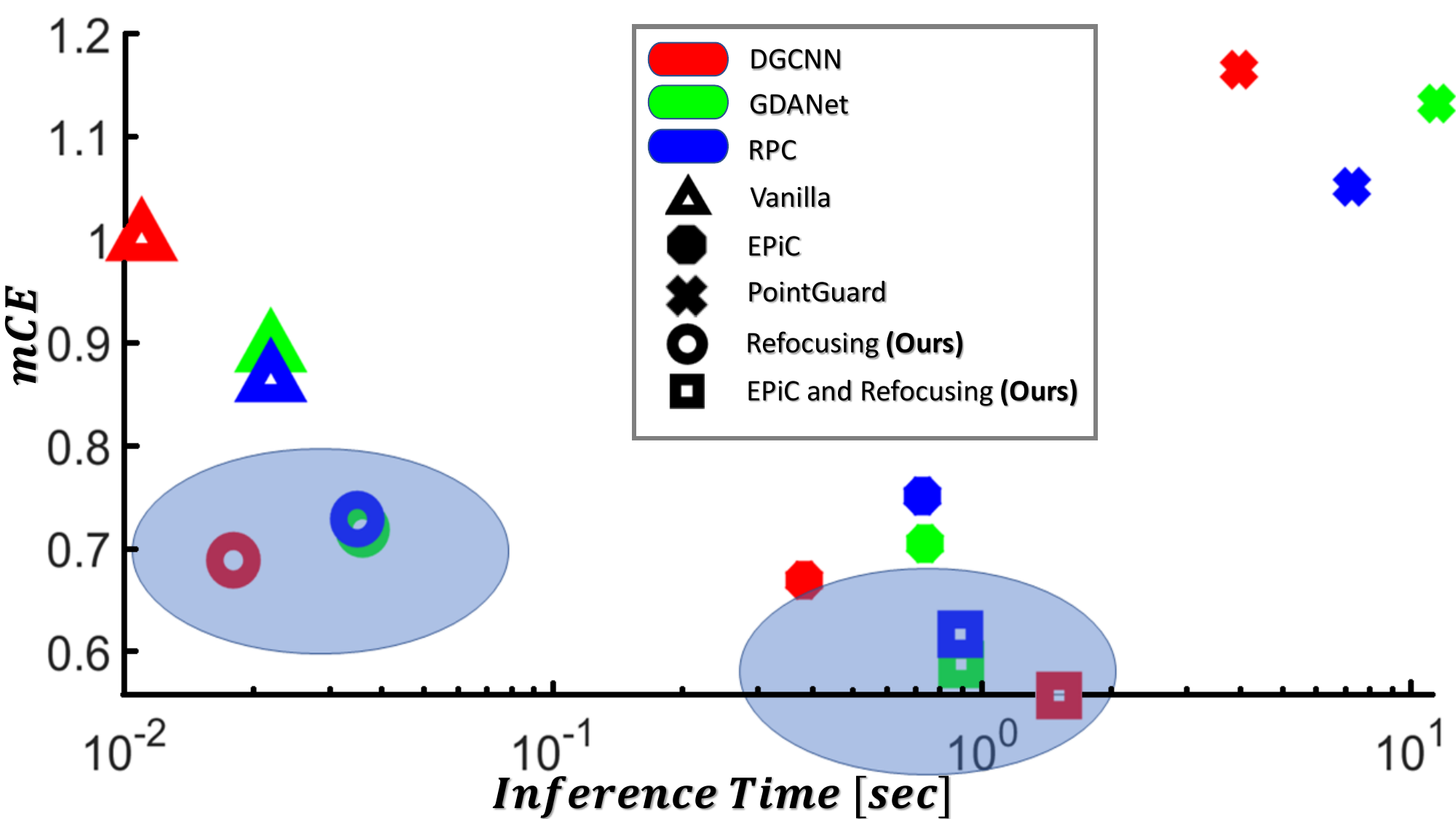}
   \caption{{\bf Robustness vs. inference time (log-scale).} Our method is as fast as vanilla networks, with outstanding robustification, on-par with EPiC \cite{epic}. Combining our approach as extra sampling strategy achieves SOTA mCE (both highlighted in blue).}
   \label{fig:timing}
\end{figure}

We evaluate various fixed values of remaining points, compare them to our adaptive threshold, and plot the accuracy and robustness in a classification task (see Fig. \ref{fig:adaptive_vs_fixed}). The results strongly support our proposal of employing an adaptive threshold rather than a fixed one. The utilization of our suggested selective filtering approach significantly improves the mean corruption error (mCE) compared to any fixed number of sampling points. Our refocusing algorithm by sampling is summarized in \cref{alg:inference} for inference; the training procedure is detailed in the supplementary.

\section{Applications and Experiments}

We now show how our proposed refocusing method can be used for robust classification and adversarial defense.


\begin{table*}
\begin{center}
  \begin{tabular}{p{2.0cm} || C{4cm} || C{1cm} C{1cm} C{1cm} C{1cm}}
  
    \hline
    \multirow{2}{*}{Model} &
    \multirow{2}{*}{Approach} &
      \multicolumn{2}{c}{WolfMix\cite{pointwolf, rsmix}}&
      \multicolumn{2}{c}{Un-Augmented}\\
      \cmidrule(l){3-4} \cmidrule(l){5-6}
    & & OA↑ & mCE↓ & OA↑ & mCE↓\\
    \hline
    \multirow{5}{*}{DGCNN\cite{dgcnn}} & Vanilla & 93.2\% & 0.590 & 92.6\% & 1.000\\
     & EPiC \cite{epic} & 92.1\% & 0.529 & 93.0\% & 0.669\\
      & PointGuard \cite{pointguard} & 81.9\% & 1.154 & 83.8\% & 1.165\\
      & Refocusing \textbf{(Ours)} & 91.4\% & 0.560 & 91.6\% & 0.688\\
     & EPiC \& Refocusing \textbf{(Ours)} & 92.9\% & \textbf{0.484} & 93.4\% & \textbf{0.557}\\
    \hline
    \multirow{5}{*}{RPC\cite{modelnet_c}} & Vanilla & 93.3\% & 0.601 & 93.0\% & 0.863\\
     & EPiC \cite{epic} & 92.7\% & 0.501 & 93.6\% & 0.750\\
      & PointGuard \cite{pointguard} & 83.2\% & 1.067 & 86.9\% & 1.051\\
      & Refocusing \textbf{(Ours)} & 91.2\% & 0.562 & 91.6\% & 0.728\\
     & EPiC \& Refocusing \textbf{(Ours)} & 92.9\% & \textbf{0.476} & 93.2\% & \textbf{0.616}\\
    \hline
    \multirow{5}{*}{GDANet\cite{gdanet}} & Vanilla & 93.4\% & 0.571 & 93.4\% & 0.892\\
     & EPiC \cite{epic} & 92.5\% & 0.530 & 93.6\% & 0.704\\
      & PointGuard \cite{pointguard} & 83.2\% & 1.059 & 84.8\% & 1.132\\
      & Refocusing \textbf{(Ours)} & 91.8\% & 0.528 & 91.4\% & 0.718\\
     & EPiC \& Refocusing \textbf{(Ours)} & 92.8\% & \textbf{0.493} & 93.4\% & \textbf{0.587}\\
    \hline
  \end{tabular}
\end{center}
\caption{\textbf{Comparison on ModelNet-C \cite{modelnet_c}, WolfMix augmented and augmented free.} Our approach is on-par with EPiC with extremely faster inference time (see Fig. \ref{fig:timing}). Combining our approach as extra sampling strategy in EPiC based on RPC achieves SOTA results in terms of robustness (lower mCE values).}
\label{table:robust_classification}
\end{table*}

\subsection{Robust classification}
\textbf{Benchmark.} ModelNet-C \cite{modelnet_c} is a variation of ModelNet-40 \cite{modelnet40} designed to assess robustness to out-of-distribution (OOD) data. It introduces seven types of corruptions (jitter, scale, rotate, add-global, add-local, drop-global, and drop-local), each with five difficulty levels. A unified calculation mechanism, referred to as \emph{mean Corruption Error (mCE)}, is used to measure robustness. Lower mCE scores indicate better performance. Please refer to \cite{modelnet_c} for more details.

During training, a point-cloud classification network is trained on the clean set only, adapted to accept a wide range of sampled points (256-1024). The same basic network is used for querying the influence map and for the actual prediction on the filtered sample. Thus, this process can be thought of as a self-restraining process. The network provides a mapping of the influential inputs. After refocusing, the same network is used for classification, based on the least influential points. 
More details and a pseudocode of the training procedure appear in the supplementary.
Inference procedure is described in details in Algorithm \ref{alg:inference}.
It includes dual forward-pass, and simple and fast extra calculations.
To demonstrate the efficiency of the proposed method, Fig. \ref{fig:timing} depicts a  comparison to other robust networks, by plotting $mCE$ vs. inference time.  We train three different networks with refocusing, showing a substantial improvement in robustness (lower $mCE$), see  \cref{table:robust_classification}. We used refocusing as an extra global sampling strategy, such that combining with EPiC \cite{epic} ensemble method, yields SOTA results on this dataset (technical explanation of embedding refocus in EPiC are provided in the supplementary).

\begin{table*}
\begin{center}
  \begin{tabular}{p{2.7cm} C{1.5cm} C{1.6cm} C{1.5cm} C{1.6cm} C{1.5cm} C{1.6cm}}
  
    \hline
    \multirow{2}{*}{Defense} &
      \multicolumn{2}{c}{DGCNN \cite{dgcnn}}&
      \multicolumn{2}{c}{PointNet\cite{pointnet}} &
      \multicolumn{2}{c}{GDANet\cite{gdanet}} \\
      \cmidrule(l){2-3} \cmidrule(l){4-5} \cmidrule(l){6-7}
    & ASR↓ & AQ↑ & ASR↓ & AQ↑ & ASR↓ & AQ↑ \\
    \hline
    Undefended & 99.3 & 106.7 & 99.8 & 18.9 & 99.8 & 18.9 \\
    SRS(50\%) \cite{dupnet}
    & 78.4 & 566.3 & 94.0 & 190.9 & 78.1 & 595.4 \\
    SRS(30\%) \cite{dupnet} & 68.6 & 790.3 & 97.6 & 93.5 & 72.4 & 714.0\\
    SOR \cite{sor} & 75.6 & 795.6 & 78.4 & 592.9 & 69.9 & 913.2 \\
    LPF-Defense\cite{lpf_defense} & 47.8 & 1148.0 & 98.2 & 123.1 & 52.6 & 1071.4\\
    Refocusing \textbf{(Ours)} & \textbf{37.5} & \textbf{1376.18} &\textbf{72.0} & \textbf{730.4} & \textbf{34.6} & \textbf{1425.5}\\
    \hline
  \end{tabular}
\end{center}
\caption{\textbf{Adversarial defenses from shape-invariant attack \cite{shape_invariant} on ModelNet40 \cite{modelnet40}.} Attack success rate (ASR, measured in percents) is consistently the lowest and mean query cost (AQ, measured in average time) is the highest, over all examined networks, compared to all other defense methods. Note that for DGCNN and GDANet ASR is extremely decreased.}
\label{table:adversarial_attack}
\end{table*}
 
\subsection{Adversarial defense}

Another major threat for point cloud classification networks is adversarial attacks. It has been shown that main classification networks are vulnerable for this attacks, even for barely distinguishable perturbed clouds \cite{shape_invariant, saliency_maps}. In these cases, the classification network has no knowledge regarding the manipulation, thus, there is a clear advantage for OOD robust approaches. We applied our method (as described in Alg. \ref{alg:inference}) as a defense scheme against Shape-Invariant Attack \cite{shape_invariant} and compared it with several OOD defenses. Our method achieves substantial improvements, reducing the \emph{attack success rate} (ASR) to a limited 37.5\% (compare to 47.8\% using LPF-Defense \cite{lpf_defense}) when embedded to DGCNN \cite{dgcnn}. We examine the case where DGCNN is functioning as both the surrogate model and the attacked model, to eliminate transferability issues. The results are shown in \cref{table:adversarial_attack}.

\section{Future Work}

The analysis of focus introduces a deeper understanding of neural network performance. It is intriguing to measure over- and under-focus characteristics in various domains, including NLP, audio, and image processing. This understanding can pave the way for a wide range of applications. In the supplementary material, we provide a very preliminary example of how facilitating refocusing or parts of the algorithm can aid in outliers removal. The idea can further extend, for instance, for developing over- or under-focus adversarial attacks, yielding specific focus values. Another potential path is guided adversarial training, which extends the focus range to the one exposed during training.


\section{Conclusion}

In this study, we introduced a novel perspective on point cloud neural network behavior through the analysis of focus. We proposed a definition for a network's focus, over-focus, and under-focus, which can be extended beyond 3D point clouds.
We observed a strong correlation between corruptions and focus distribution. The presence of outliers predominantly increases focus, while occluded parts have the opposite effect. To enhance the network's ability to process corrupted data, we proposed a robust algorithm aimed at screening out the most influential input elements. Filtering mitigates the impact of outliers and aligns the focus distribution, resulting in improved robustness against OOD corruptions, with only a marginal degradation in accuracy for clean data.

Our method is computationally efficient, making it applicable to time-demanding applications. Experimental results on robust classification and adversarial defense tasks showcase the effectiveness of our approach. We achieved state-of-the-art results for both robust zero-shot classification on the ModelNet-C \cite{modelnet_c} dataset and for adversarial defense against Shape-Invariant attacks \cite{shape_invariant}.

\clearpage  

%
%
\bibliographystyle{splncs04}
\bibliography{main}

\pagebreak
\section{Filtering in feature space}
The algorithm outlined in the main paper involves a dual forward pass. The first pass is necessary for querying the importance score, while the second pass is utilized to predict the point cloud filtered based on the previously calculated importance score. Employing filtering in Euclidean space serves as a safeguard against the propagation of outliers from the primal space to the feature space.
An alternative approach we propose is to perform filtering directly in the feature space instead of the input space. This alternative method requires only a single forward pass of the base network, incurring no additional latency. This approach can be likened to a dropout layer that emphasizes the filtration of dominant feature elements instead of random ones.
While feature space filtering yields lower robustness compared to filtering in the primal space, as outliers in the Euclidean space can still contaminate features, it does provide a notable boost in robustness, compared to the vanilla base network.
Thus, if speed is essential, one can consider also this approach in point cloud classification networks. 
To illustrate this improvement, we integrated feature filtering into the DGCNN and RPC models and evaluated their performance on the ModelNet-C dataset. The results, as presented in Tab. \ref{table:critical_points_dropout}, highlight the significant enhancement in robustness achieved through the implementation of our proposed layer. This improvement is demonstrated in comparison to the vanilla version of the network, with no associated latency cost.

\begin{table}
\begin{center}
  \begin{tabular}{p{1.2cm} || C{6.0cm} || C{0.8cm} C{0.8cm} C{2.5cm}}
  
    \hline
    Model & Approach & OA\% & mCE & \#Forward Pass\\
    \hline
    \multirow{1}{*}{DGCNN} & Vanilla & 92.6 & 1.000 & 1\\
    & Refocusing feature space.\textbf{(Ours)} & 92.7 & 0.818 & 1\\
     & Refocusing Euclidean space.\textbf{(Ours)} & 91.6 & 0.688 & 2\\
    \hline
    \multirow{1}{*}{RPC} & Vanilla & 93.0 & 0.863 & 1\\
     & Refocusing feature space\textbf{(Ours)} & 92.0 & 0.797 & 1\\
     & Refocusing Euclidean space \textbf{(Ours)} & 91.6 & 0.728 & 2\\
    \hline
  \end{tabular}
\end{center}
\caption{\textbf{Filtering in feature space vs. Euclidean space using Refocusing} Employing \textit{Refocusing} for filtering in the feature space, as opposed to the primary Euclidean space, enhances robustness compared to the standard network, all without incurring any latency costs.}
\label{table:critical_points_dropout}
\end{table}

\section{Benchmarks}
\textbf{ModelNet40}.
Synthetic dataset, constitutes a widely employed collection of CAD meshes spanning 40 distinct classes, encompassing objects such as monitors, beds and persons. These meshes have undergone uniform sampling procedures, resulting in the formation of 3D point clouds, each comprising 1024 points. The dataset encompasses a total of 12,311 samples, partitioned into 9,843 samples designated for training and 2,468 samples allocated for testing purposes.

\noindent \textbf{ModelNet-C}. Ren et al., introduced a corrupted point cloud benchmark denoted as ModelNet-C, based on ModelNet40, to facilitate the OOD robustness. ModelNet-C encompasses a spectrum of seven distinct corruption types, namely jitter, scale, rotation, add-global, add-local, drop-global, and drop-local, each exhibiting five levels of difficulty. To quantitatively gauge robustness, introduced a comprehensive metric termed mean Corruption Error (mCE), a relative robustness measure with DGCNN algorithm serving as a pivot network (and thus by definition mCE value of 1). Given the error-based nature of this metric, a lower mCE score is indicative of superior performance. 

\noindent \textbf{ScanObjectNN}. Contains 2902 point clouds spanning 15 distinct categories, constituting a more intricate collection that arises from real-world scans featuring background elements and instances of occlusion. There exist classes which overlapp with ModelNet40 classes e.g. chairs, desks and sofas.

\section{Implementation details}
The procedure of the training phase is provided in Alg. 
\ref{alg:training}.
\begin{algorithm}
\caption{$Refocusing$ (Training)}\label{alg:training}
\begin{algorithmic}
\Require{$DataSet, Influence$} 
\For {$X, label \in DataSet$}
\State $X_f = model(X)$ 
\State $\hat{I}_F(X) = \frac{Influence(X)}{\Sigma_{i=1}^N(Influence(X_i))}$ 
\State $H_{n} = NormalizedEntropy(\hat{I}_F)$
\State $K = rand(256,1024)$  
\State $X_{sampled} = SelectLowest(X, \hat{I}_F, K)$
\State $P = model(X_{sampled})$ 
\State $L = CrossEntropyLoss(P,label)$ 
\State $L.backward()$ 
\EndFor
\State $params_{Refocusing} \gets BestModelParams$
\end{algorithmic}
\end{algorithm}


\subsection{Adversarial defense}
\textbf{Defense parameters.} Following the Shape-Invariant method, SRS was applied with 30\% of filtered points, as well as with 50\%, while SOR was applied with 2 KNN and $\sigma=1.1$.
Regarding LPF-Defense, we utilized $l_{max}=16$, and in line with the ablation study conducted in LPF-Defense paper, we executed it with $\sigma=20$. The paper recommends training the network on the clean dataset and employing these parameters during inference. Consequently, for each network, training was performed with these parameters and  testing was conducted using the same parameters. The training process was over 300 epochs, employing the ADAM optimizer with a learning rate of $5^{-3}$, momentum of $0.9$, and weight decay of $1^{-4}$.

\noindent \textbf{Attack setting.} To underscore our defense capabilities, and recognizing that the Shape-Invariant attack leverages a sensitivity map characterized by transferability across diverse networks, we decided to examine the most challenging scenario. This entails investigating the situation where the surrogate model, responsible for generating the sensitivity map, aligns with the victim model. DGCNN serves as both the surrogate model and the victim. Similarly, PointNet fulfills dual roles as the surrogate and victim, as does GDANet.
\subsection{Robust classification}
In the unaugmented setting, we initially apply conventional augmentation methodologies to adhere to the OOD principle. In contrast, for the augmented iteration using WolfMix, the augmentation process is first applied to the sample, followed by our importance-based subsampling technique. To mitigate the effects of randomness, a predetermined seed is utilized.
All models undergo training for 300 epochs in the unaugmented scenario and 500 epochs in the augmented case, utilizing a learning rate of 5e-4. A cosine annealing scheduler is employed to drive the learning rate to converge to zero. A batch size of 64 is adopted.
In the unaugmented version, the augmentation protocol outlined by DGCNN is followed, encompassing two steps: 1) stochastic anisotropic scaling spanning the range of [2/3, 3/2]; and 2) random translation within the interval of [-0.2, +0.2]. The implementation makes use of the PyTorch library. The primary training objective involves minimizing the Cross-Entropy loss.
During training, we computed the importance score and cropped the sample to a random size within the range of [256, 1024] contains points with the lowest importance. This prepares the model to accommodate a wide range of sample sizes during inference, where we crop the sample to an unknown size based on the adaptive threshold. Due to variations in the number of points in each cloud, inference is performed using a batch size of 1, a common practice in real-world applications. Timing estimates are obtained by averaging the time taken for 100 iterations of batches with a size of 1.

The combination of our sampling approach with EPiC is achieved by applying our sub-sampling technique once for each sample, resulting in a fixed size of 600 points. The prediction is then duplicated four times and concatenated with other sub-samples. This duplication is carried out to ensure an equal impact for each of the sampling schemes. Consequently, the ensemble created using our sampling approach comprises 16 members.

\section{Full results}
We present comprehensive and detailed results encompassing both the adversarial attack experiment (refer to Tab. \ref{table:adversarial_attack_supp}) and the robust classification analysis (refer to Tab. \ref{table:robust_classification_full}). 
\textbf{Adversarial attack experiment.} The comprehensive table includes three additional distance measures between adversarial and benign shapes: Chamfer, Hausdorff, and MSE distances.
\textbf{Robust classification}. The detailed table contains additional results for mCE under each specified corruption. Notably, certain defenses display a correlation with specific corruptions across various networks. For instance, Drop-Local is most effectively mitigated by EPiC, while scale is best addressed by PointGuard. However, when considering the mCE metric, which aggregates the overall robustness scores across all corruptions, EPiC \& Refocusing (our method combined with EPiC) consistently achieve the highest scores among the networks examined.



\begin{table*}
\begin{center}
\begin{tabular}{c || c || c c c c c}
\hline
Surrogate & Defense & ASR(\%) & A.Q(times) & C.D($10^{-4}$) & H.D($10^{-2}$) & MSE \\
\hline\hline
DGCNN & -- & 99.3 & 106.7 & 3.18 & 4.54 & 1.22\\
& SOR & 75.6 & 795.6 & 2.59 & 3.48 & 1.65\\
& SRS (50\%) & 78.4 & 566.3 & 1.21 & 3.59 & 0.69\\
& SRS (30\%) & 68.6 & 790.3 & 1.64 & 4.11 & 0.85\\
& LPF-Defense & 47.8 & 1148.0 & 1.84 & 4.09 & 0.93\\
& Refocusing (Ours) - Fixed (600) & 43.5 & 1265.0 & 2.30 & 4.16 & 1.11\\
& Refocusing (Ours) - Adaptive& \textbf{37.5} & \textbf{1376.1} & 2.28 & 4.06 & 1.17\\
\hline
PointNet  & - & 99.8 & 18.9 & 2.09 & 4.46 & 0.80\\
& SOR  & 78.4 & 592.9 & 9.26 & 4.33 & 2.74\\
& SRS (50\%) & 94.0 & 190.9 & 1.67 & 4.42 & 0.71\\
& SRS (30\%) & 97.6 & 93.5 & 1.76 & 4.43 & 0.72\\
& LPF-Defense & 98.2 & 123.1 & 5.36 & 4.56 & 1.41\\
& Refocusing (Ours) - Fixed (600) & 74.1 & 693.6 & 11.33 & 4.70 & 2.66\\
& Refocusing (Ours) - Adaptive & \textbf{72.0} & \textbf{730.4} & 11.13 & 4.68 & 2.72\\
\hline
GDANet  & - & 99.4 & 95.6 & 4.16 & 4.67 & 1.27 \\
& SOR  & 69.9 & 913.2 & 3.67 & 3.98 & 1.78\\
& SRS (50\%) & 78.1 & 595.4 & 1.70 & 4.16 & 0.78\\
& SRS (30\%) & 72.4 & 714.0 & 2.19 & 4.42 & 0.92\\
& LPF-Defense & 52.6 & 1071.48 & 2.18 & 4.26 & 0.94\\
& Refocusing (Ours) - Fixed (600) & \textbf{32.9} & \textbf{1447.3} & 2.18 & 4.32 & 1.03\\
& Refocusing (Ours) - Adaptive & 34.6 & 1425.5 & 2.13 & 4.21 & 1.03\\
\hline
\end{tabular}
\end{center}
\caption{\textbf{Comprehensive table of adversarial defenses against Shape-Invariant attack on ModelNet40.} Incorporating our approach with a fixed threshold (600 points). Our adaptive approach outperforms the fixed threshold in two out of the three evaluated networks.}
\label{table:adversarial_attack_supp}
\end{table*}

\begin{table*}
\begin{center}
  \begin{tabular}{p{1.2cm} || C{2.5cm} || C{0.78cm} C{0.8cm} C{0.9cm} C{0.9cm} C{0.9cm} C{0.9cm} C{0.9cm} C{0.9cm} C{0.9cm}}
  
    \hline
    Model & Approach & OA\% & mCE & Scale & Jitter & DropGlobal & Drop-Local & Add-Global & Add-Local & Rotate\\
    \hline
    \multirow{5}{*}{DGCNN} & Vanilla & 92.6 & 1.000 & 1.000 & 1.000 & 1.000 & 1.000 & 1.000 & 1.000 & 1.000\\
     & EPiC & 93.0 & 0.669 & 1.000 & 0.680 & \underline{0.331} & \underline{0.498} & 0.349 & 0.807 & 1.019\\
      & PointGuard  & 83.8 & 1.165 & \underline{0.743} & 0.809 & 0.840 & 0.812 & 0.838 & 0.785 & \underline{0.584}\\
      & Refocusing(Ours) & 91.6 & 0.688 & 0.896 & 0.868 & 0.891 & 0.820 & 0.902 & 0.829 & 0.779\\
     & EPiC \& Refocusing (Ours) & \underline{93.4} & \underline{0.557} & 0.957 & \underline{0.494} & 0.335 & 0.522 & \underline{0.258} & \underline{0.382} & 0.949\\
    \hline
    \multirow{5}{*}{RPC} & Vanilla & 93.0 & 0.863 & 0.840 & 0.892 & 0.492 & 0.797 & 0.929 & 1.011 & 1.079\\
     & EPiC & 93.6 & 0.750 & 0.915 & 1.057 & 0.323 & \underline{0.440} & 0.281 & 0.902 & 1.330\\
      & PointGuard & 86.9 & 1.051 & \underline{0.773} & 0.817 & 0.868 & 0.843 & 0.867 & 0.804 & \underline{0.590}\\
      & Refocusing(Ours) & 91.6 & 0.728 & 0.901 & 0.808 & 0.912 & 0.842 & 0.862 & 0.819 & 0.744\\
     & EPiC \& Refocusing(Ours) & \underline{93.2} & \underline{0.616} & 0.957 & \underline{0.690} & \underline{0.319} & 0.464 & \underline{0.258} & \underline{0.444} & 1.177\\
    \hline
    \multirow{5}{*}{GDANet} & Vanilla & 93.4 & 0.892 & 0.830 & 0.839 & 0.794 & 0.894 & 0.871 & 1.036 & 0.981\\
     & EPiC & \underline{93.6} & 0.704 & 0.936 & 0.864 & \underline{0.315} & \underline{0.478} & 0.295 & 0.862 & 1.177\\
      & PointGuard & 84.8 & 1.132 & \underline{0.755} & 0.804 & 0.847 & 0.819 & 0.846 & 0.787 & \underline{0.589}\\
      & Refocusing(Ours) & 91.4 & 0.718 & 0.900 & 0.848 & 0.884 & 0.812 & 0.908 & 0.810 & 0.763\\
     & EPiC \& Refocusing(Ours) & 93.4 & \underline{0.587} & 0.926 & \underline{0.617} & 0.323 & 0.512 & \underline{0.258} & \underline{0.393} & 1.079\\
    \hline
  \end{tabular}
\end{center}
\caption{\textbf{Comprehensive comparison table for augmented free ModelNet-C.}}
\label{table:robust_classification_full}
\end{table*}

\section{Discussion and future directions}

\begin{figure*}[ptbh!]
  \centering
   \includegraphics[width = 0.8\linewidth, height =0.332
   \linewidth]{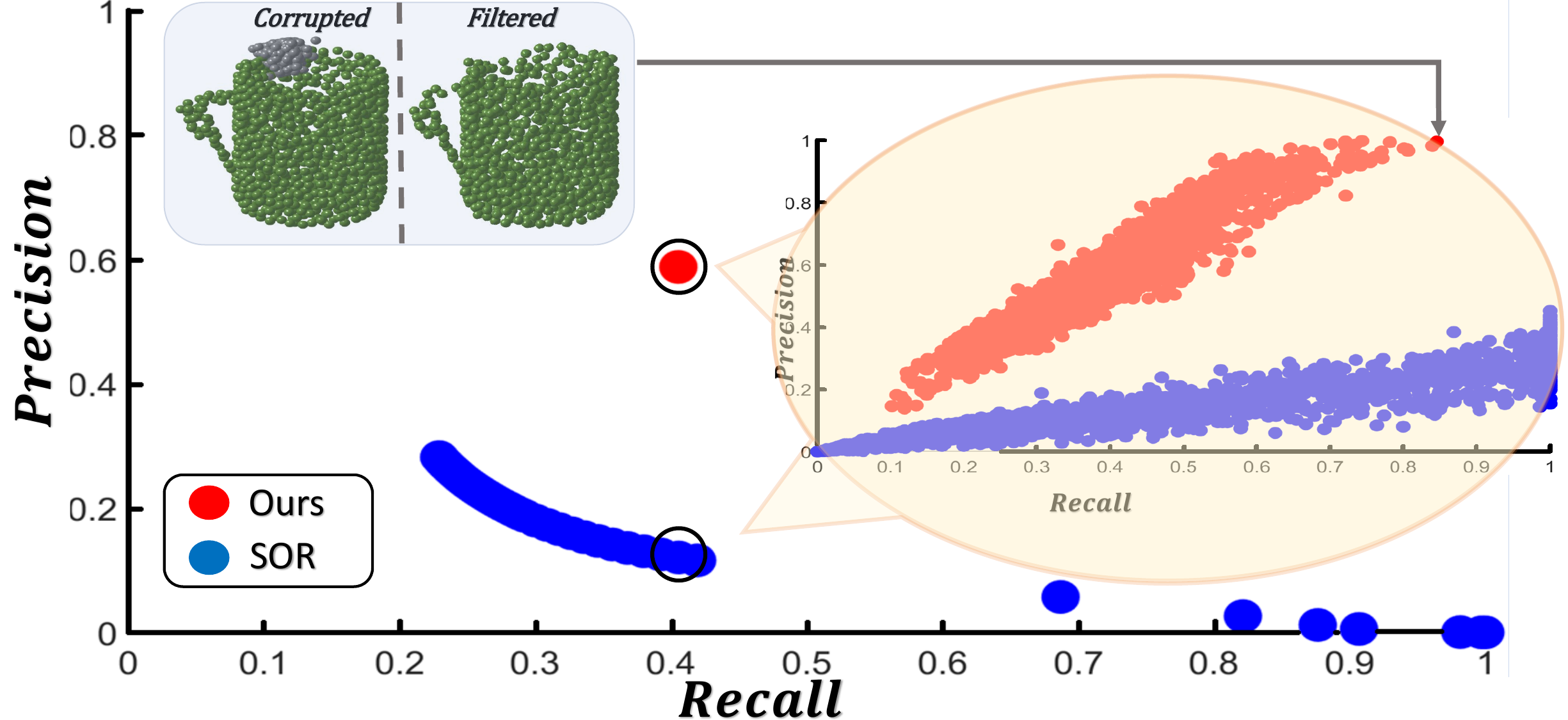}
   \caption{{\bf Outlier removal on add-local.} Refocusing outperforms SOR investigated on a grid of standard deviations. {\bf Zoom-in on SOR with corresponding recall.} Our approach achieves superior results on each sample, precise removal is exemplified by the cup shapes.}

   \label{fig:outlier_removal}
\end{figure*}

\noindent {\bf Outliers removal.}
In outlier detection and removal, two main categories of methods are prevalent. Learnable methods, which
require clean and corrupted pairs as ground truth, thus unsuitable for the OOD regime, which is our primary focus. Classical methods are a valid choice, with statistical outlier removal (SOR) being a common example. We note that SOR requires prior knowledge on the outlier characteristics. Moreover, its  performance deteriorates for subtle, smooth corruptions, found in real-world scenarios.
For outliers removal task we propose to use a another variant of influence, defined as: $I_F(i) = \sum_{k=1}^G |X_G(i,k)|.$
And average as adaptive threshold as follows: $S_X := \{X_i \, :\, I_F(i) \leq \frac{\Sigma_{i=1}^{N}(I_F(i))}{N}\}$.
A preliminary evaluation of our approach appears promising. We attempt to detect Add-local corruption of ModelNet-C. For SOR, we experiment with various $\sigma$ values, calculating mean recall and precision for each. We focus on the settings that yield the same recall as our method and present recall-precision values per sample (See Fig. \ref{fig:outlier_removal}). Our findings conclusively demonstrate the superior performance of our approach in effective outlier detection.

\begin{figure}[ptbh!]
  \centering
   \includegraphics[width = 0.8\linewidth, height =0.48 
   \linewidth]{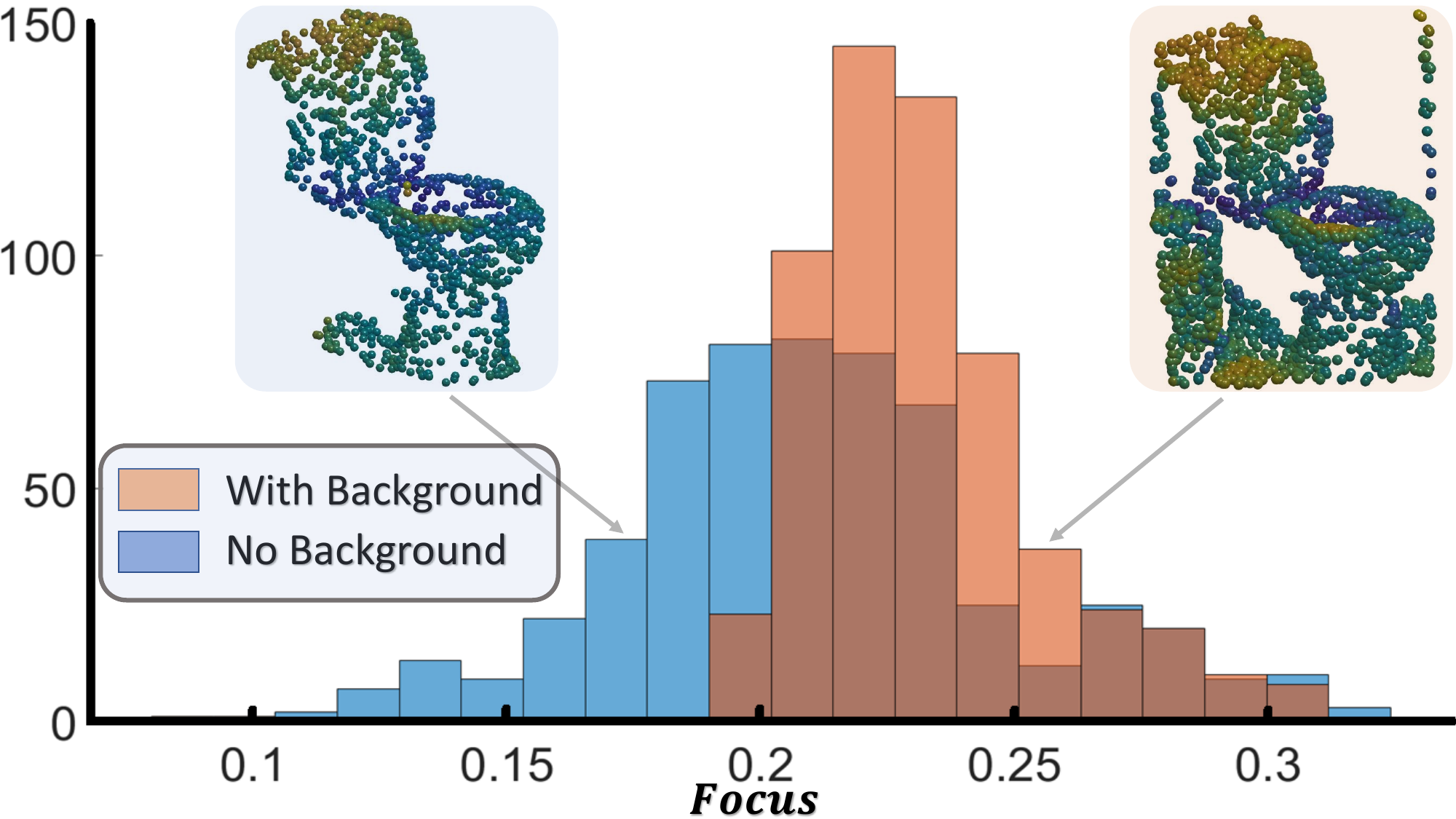}
   \caption{{\bf Focus histogram on ScanObjectNN.} Samples contains background yield higher focus than those without background.}
   \label{fig:scan_object_nn_success_rate}
\end{figure}

\noindent {\bf Statistical acquisition analysis.}
One can analyze certain statistical aspects of different 
acquisition scenarios. For instance, acquisition of objects containing background and ones which do not. Samples with background exhibit relatively higher focus, as the background may contain prominent features that divert the network's attention. See Fig. \ref{fig:scan_object_nn_success_rate} for  analysis of the real-world ScanObjectNN dataset.

\end{document}